# Vertical Standardisation for High-Risk AI Systems under the EU AI Act: A Domain-Specific Framework for Algorithmic Hiring

*Anna Gatzioura[1], Vrettos Moulos[2] and Nina Baranowska[3]*

***Abstract***

According to the recent European legislation, high-risk AI systems will have to adapt in order to comply with requirements related to specific areas, like risk management, data quality and governance, logging and traceability, technical documentation, transparency, human oversight, and accuracy, as outlined in the European Artificial Intelligence (AI) Act.

As the standardisation process for AI is expected to remain iterative and, so far, there are no European standards on AI fully covering the challenges of algorithmic hiring, we propose specific standardisation-oriented recommendations related to the relevant AI areas specified by the European Commission. For each of these areas, we set the context by describing the requirements that AI systems in high-risk domains, and especially in recruitment, should fulfil, as well as the activities that should be carried out to ensure their appropriate use and desired performance, in line with the requirements deriving from the AI Act.

Unlike existing horizontal approaches to AI governance and standardisation, this paper contributes a vertical, domain-specific framework for algorithmic hiring, and especially ranking-based recruitment systems, by mapping the requirements of the AI Act to concrete standardisation recommendations, focusing on lifecycle discrimination risks, fairness-aware data governance, explainability, human oversight, and post-deployment monitoring in recruitment systems. Even though our recommendations were informed by the outcomes of the European project FINDHR, they are not tied to the project's technical artefacts and could be implemented using alternative methods, tools, or governance mechanisms.

---

[1] University Pompeu Fabra, Barcelona, Spain, anna.gkatzioura@upf.edu

[2] University of Piraeus, Piraeus, Greece, vrettos@unipi.gr

[3] Leiden University, Leiden, Netherlands, n.n.baranowska@law.leidenuniv.nl

## 1. Introduction

AI is becoming increasingly integrated into software development processes, while its use is expanding horizontally across a wide range of application domains. According to the EU Artificial Intelligence Act (AI Act)[4], an AI system is considered high-risk if it falls into one of two categories: (i) it is used as a safety component or as a product regulated by EU legislation listed in Annex I and is subject to a third-party conformity assessment under that legislation; (ii) it is used in one of the specific high-risk areas listed in Annex III, where it could significantly affect individuals' health, safety, or fundamental rights; these areas include, among others, employment, worker management, and access to self-employment[5]. In particular, AI systems intended to be used for the recruitment or selection of natural persons, in particular to place targeted job advertisements, to analyse and filter job applications, and to evaluate candidates, are considered as high-risk[6].

Following the AI Act, high-risk AI systems will gradually[7] need to adapt in order to comply with requirements related to risk management, data quality and governance, logging and traceability, technical documentation, transparency, human oversight, accuracy, robustness, and cybersecurity as outlined in the AI Act (mainly in Articles 9 to 15 as well as Articles 43, 72 and 73). Compliance with these requirements will be ensured through the quality management systems and conformity assessments before and after placing the system on the market. In this context, standardisation plays a key role as it enables AI providers to demonstrate compliance with the AI Act, since conformity with harmonised standards, reflecting the state of the art, creates a presumption of conformity with the Act's requirements[8]. Currently, the standardisation process for AI, run by standardisation bodies, like CEN-CENELEC, is still ongoing, and so far, there are no EU standards on AI fully covering algorithmic hiring challenges[9].

To advance technical harmonisation and to prepare the necessary technical environment for the implementation of the AI Act in the area of algorithmic hiring, we propose multiple contributions to vertical harmonised standards for AI Act compliance, aligned with the areas identified in the European Commission's standardisation request. We base our recommendations on the outcomes of the European project FINDHR[10] (Fairness and Intersectional Non-Discrimination in Human Recommendation) that focused on algorithmic hiring, understood as the use of AI for human recommendation, and particularly centred on AI-enabled ranking systems used to support recruitment processes by ordering candidates based on their estimated suitability for a given job position. Therefore, based on the developed materials (anti-discrimination methods, algorithms, and training materials for AI) and the designed best practices, in this standardisation-oriented contribution we propose standardisation actions to address the specific aspects that AI systems for human recommendation should address, to detect and mitigate discrimination along the multiple stages of the recruitment pipeline, and how, in order to comply with the AI Act. The proposed recommendations focus on two directions in the recruitment industry: algorithmic auditing and software development, with the aim to cover all the stages of the lifecycle of an AI recruitment system, as well as the multiple stakeholders involved.

The remainder of the paper is structured as follows: in section 2 we present the current standardisation context, with focus on the European landscape and especially in relation to the AI Act. We highlight the key characteristics that are expected of harmonised standards under the AI Act, as well as of the systems developed in line with those. Furthermore, we describe the concrete areas of AI that standardisation deliverables should address, as identified by the European Commission. Finally we introduce our specific case study: the FINDHR European project. In section 3 we present in detail the developed specific recommendations, to address AI Act compliant vertical standardisation actions in the algorithmic hiring domain. We conclude our work in section 4 by discussing the implications of the proposed recommendations and their relevance to forthcoming regulatory and standardisation developments.

---

[4] Regulation (EU) 2024/1689 of the European Parliament and of the Council of 13 June 2024 laying down harmonised rules on artificial intelligence (Artificial Intelligence Act), OJ L, 2024/1689, http://data.europa.eu/eli/reg/2024/1689/oj.

[5] AI Act, Article 6.

[6] AI Act, Annex III point 4(a)

[7] See the summary of key dates: https://artificialintelligenceact.eu/implementation-timeline/

[8] Recital 121 and the Article 43 of the AI Act

[9] https://www.euronews.com/next/2025/04/16/eu-standards-bodies-flag-delays-to-work-on-ai-act

[10] https://findhr.eu/

## 2. Current standardisation context

### 2.1. Existing standardisation bodies and processes

For the purpose of generating strong, internationally relevant anti-discrimination methodologies in algorithmic hiring, a large and wide array of AI standardisation initiatives and policy frameworks have been studied. We divided those into two categories, the **Frameworks & Principles** and the **Technical Standards**.

Relevant global references in the first category include the NIST AI Risk Management Framework[11], the OECD AI Principles[12], the Blueprint for an AI Bill of Rights (U.S.)[13], UNESCO's Recommendation on the Ethics of AI[14], and the Council of Europe Framework Convention on AI, Human Rights, Democracy, and the Rule of Law[15]. Each of them touches on vital elements of an AI system development and deployment —from ethical protections and human rights to risk management — throughout the systems' lifecycle and to organisational governance. The second category comprises technical standards such as ISO/IEC 42001 (AI Management Systems)[16], ISO/IEC 23894 (AI Risk Management)[17], and IEEE standards, including P2863[18] and P7003[19,20], serving as technical references in responsible AI development and the mitigation of bias in automated decision systems.

We have recognised a recurring challenge in the standardisation of AI, particularly regarding the protection of fundamental rights and the complexity of the hiring processes[21]. Traditionally, technical standards have not addressed social or ethical impacts, nor fundamental rights, which are now central to the regulation of AI[22]. Instead, standards have historically focused on technical requirements, product safety, and market harmonisation, particularly in sectors such as automotive, electronics, and manufacturing. These standards were largely developed by technical and industry experts and centred on measurable specifications.

The emergence of AI introduces a new kind of challenge. AI systems can pose significant risks to fundamental rights, such as the right to non-discrimination. As the standardisation process remains largely technical, it raises important questions about whether tech and industry stakeholders alone should define standards that have wide-reaching implications for fundamental rights. Although Article 40 of the AI Act emphasises on the importance of multi-stakeholder governance in the standardisation process, including balanced representation of interests and the effective participation of all relevant stakeholders, the actual involvement of civil society, SMEs, and researchers may nevertheless remain difficult to achieve in practice[23].

The hiring process adds another layer of complexity. Algorithmic hiring is not a single point of decision-making, but rather a series of interdependent stages, from job advertising and CV screening

---

[11] Artificial Intelligence Risk Management Framework (AI RMF 1.0)

[12] AI principles | OECD

[13] Blueprint for an AI Bill of Rights | OSTP | The White House

[14] Recommendation on the Ethics of Artificial Intelligence - UNESCO Digital Library

[15] The Framework Convention on Artificial Intelligence - Artificial Intelligence

[16] ISO/IEC 42001:2023 - AI management systems

[17] ISO/IEC 23894:2023 - AI — Guidance on risk management

[18] IEEE SA - P2863

[19] IEEE SA - IEEE 7003-2024

[20] IEEE P7003TM Standard for Algorithmic Bias Considerations | IEEE Conference Publication | IEEE Xplore

[21] Leyden, A. (2025). Standards and the EU AI act: legitimacy, state of play, and future challenges. Information & Communications Technology Law, 1–31. https://doi.org/10.1080/13600834.2025.2570966

[22] https://policyreview.info/articles/analysis/regulating-ai-through-technical-standards

[23] https://edri.org/wp-content/uploads/2022/05/The-role-of-standards-and-standardisation-processes-in-the-EUs-Artificial-Intelligence-AI-Act.pdf

to interviews, assessments, and final selection. Each of these stages can amplify discriminatory bias if not carefully designed and monitored, therefore a holistic approach, addressing the entire AI lifecycle, is required to fight against discrimination in AI recruitment systems, from the design of fair, interpretable, and bias-aware systems to ongoing monitoring. This approach should be grounded in multidisciplinary collaboration, bringing together legal, ethical, social, and technical parties.

With all this in mind, to contribute to the international standardisation discourse we provide field-specific recommendations that address the complexity of algorithmic hiring, in alignment with wider frameworks, including Singapore's AI Verify and Generative AI Governance Framework[24], the G7 Hiroshima Process Guiding Principles[25], and the EU-oriented CEN-CENELEC' s efforts[26] towards the AI Act.

From the Frameworks & Principles perspective, the main focus has been on the guidance and policy direction to ensure that AI technologies relate to society. Given the complexity and layered nature of hiring processes, only a complete, end-to-end approach could contribute to fairness and non-discrimination. The NIST AI Risk Management Framework centers on trustworthiness, providing a structured guidance to map, measure, and manage risk in AI systems. The Blueprint for an AI Bill of Rights (U.S.) encompasses five central rights to be protected for individuals dealing with automated systems, including protections against algorithmic discrimination and notice requirements. OECD AI Principles lay the groundwork for designing human-centred AI with emphasis on transparency and accountability. The Council of Europe's framework convention places strong weight on the protection of fundamental rights and democratic values. The Singapore AI Verify Framework emphasises on the testing and validation of trustworthy AI systems, while the Singapore Generative AI Governance Framework takes this vision further towards generative models. Finally, the UNESCO AI Ethics Recommendation stands as the global compass for the ethical use of AI, while the G7 Hiroshima Process Guiding Principles stand for high-level commitment toward the development of safe and human-centric AI. These frameworks, therefore, provide different and complementary perspectives, necessary to ensure human rights, transparency, and fairness in all stages of algorithmic hiring pipelines.

From the International Standards perspective, a strong emphasis is placed on how various technical and procedural controls are applied to AI systems, including algorithmic hiring. These standards are not checklists for mere compliance, they project expectations as to the manner in which AI systems are designed, developed, validated, and monitored throughout their lifecycle. For instance, ISO/IEC 42001 specifies an AI management system that an organisation may adopt to instil governance and risk-aware culture into their processes. ISO/IEC 23894 gives direction on AI-specific risk management so that developers can recognise, treat, and reduce risks during deployment of the system. Various initiatives from IEEE strengthen this approach: P2863 focuses on organisational governance of AI, emphasising accountability and alignment to ethical principles; while P7003 addresses algorithmic bias, providing methods for identifying, and reducing, discriminatory impacts.

## 2.2. European standardisation under the AI Act

The AI Act represents one of the first comprehensive attempts to set legal rules for the safe development and deployment of AI systems. It has been inspired by the New Legislative Framework (NLF), under which EU law does not directly deal with technical specifications but sets out essential "requirements" that systems must meet, which leaves some flexibility on the means of compliance[27]. Based on Recital 121 of the AI Act, which stresses that standardisation should play a key role in providing technical solutions to ensure compliance, providers of AI systems will be strongly incentivised to rely on harmonised standards, since the conformity with such standards gives presumption of conformity with the AI Act according to Article 40. Therefore, a set of harmonised standards, translating legal requirements into technical characteristics are crucial in practice. As of now, the number of standards focusing on AI applications, and especially of high-risk, is still very limited coming from the ongoing work of standardisation bodies, like IEEE AI Standards Committee, ETSI and CEN-CENELEC JTC 21.

[24] Proposed_MGF_Gen_AI_2024.pdf

[25] Hiroshima Process International Guiding Principles for Advanced AI system | Shaping Europe's digital future

[26] European AI Standardisation | CEN-CENELEC JTC 21 |

[27] https://www.tandfonline.com/doi/full/10.1080/13600834.2025.2570966#d1e561

We believe that, in order to advance technical harmonisation in the field of AI and to prepare the necessary technical environment for the implementation of the AI Act, it is important to advance the work on European standards, especially in relation to high-risk applications of AI, to support the key elements defined in the AI Act.

Several key characteristics are expected of harmonised standards under the AI Act, as well as of the systems developed in line with these standards[28, 29] including that they should be:

- Tailored to the **objectives of the AI Act**: Existing international standardisation efforts tend to focus on protecting the objectives of organisations using AI. By contrast, the AI Act redefines the roles of standards by integrating the protection of fundamental rights of individuals, alongside the promotion of innovation, competitiveness, and growth in the internal market.
- **Oriented to AI systems and products**: The developed standards should cover all the phases of an AI product lifecycle, from the initial inception to the post-market monitoring.
- **Prescriptive and clear**: The standards should provide in explicit terms the processes, techniques and methods that need to be in place to make AI systems trustworthy, while being mindful of the implementation complexities.
- Aligned with the **state of the art**: The techniques used in modern AI systems should be considered, their architectures, their predicted evolution and the disruptions they may cause (for instance: generative AI).
- Applicable **across sectors and systems**: to the extent possible, the requirements defined in standards should be applicable horizontally, to various types of systems, and potentially complemented with sector specific requirements.
- **Cohesive and complementary**: the different standards proposed for the AI Act should be complementary to ensure that the maximum possible aspects are covered. A small number will focus on the core horizontal requirements while the rest will provide more specific implementation details for concrete domains and types of systems.

Furthermore, the European Commission has identified ten (10) concrete **areas** of AI that standardisation deliverables should address. These areas are summarised below[30, 31]:

- **Risk Management**: strong assurance should be provided that all the relevant foreseeable risks of an AI system have been identified and addressed. Additionally, the effectiveness of the risk mitigation measures put in place should be demonstrated through testing and evaluation based on suitable processes, metrics and thresholds. Clear and explicit requirements on the processes and outcomes to be achieved, as well as key criteria and priorities that providers of AI systems must observe, e.g. when testing mitigation measures, should be defined.
- **Data Governance and Quality**: the adoption of data-related measures, specifically tailored to the previously identified risks. In addition to the definition of explicit criteria for the selection of data quality metrics and governance processes, standards should define the evidence required to support these choices, i.e. to demonstrate their suitability and effectiveness in complying with legal obligations. At the technical level, standards should address some of the unique aspects of data-driven AI systems, e.g. those based on machine learning. They should also consider the increasing complexity of data provenance in modern AI systems, and the often-unexpected ways in which data quality issues result in downstream risks when AI systems are in operation.

---

[28] Soler Garrido, J., De Nigris, S., Bassani, E., Sanchez, I., Evas, T., André, A. and Boulangé, T., Harmonised Standards for the European AI Act

[29] Mueck, M., Forbes, R., Cadzow, S., Wood, S., Gazis, E., Gossard, C., Ortolan, F., Frost, L., Farhadi, H., Schneider, M. Böcker, M., ETSI Activities in the field of Artificial Intelligence Preparing the implementation of the European AI Act, ISBN: 979108262073

[30] COMMISSION IMPLEMENTING DECISION on a standardisation request to the European Committee for Standardisation and the European Committee for Electrotechnical Standardisation as regards high-risk AI-systems in support of Regulation (EU) 2024/1689 of the European Parliament and of the Council and repealing Implementing Decision C(2023)3215 https://ec.europa.eu/transparency/documents-register/detail?ref=C(2025)3871&lang=en

[31] Soler Garrido, J., Fano Yela, D., Panigutti, C., Junklewitz, H., Hamon, R., Evas, T., André, A. and Scalzo, S., Analysis of the preliminary AI standardisation work plan in support of the AI Act. JRC132833

- **Record Keeping**: the identification of situations that may result in risks, and the recording of all operation and performance aspects of AI systems needed to monitor compliance with the full set of legal requirements, including after being placed on the market and deployed in operation. Standards must ensure logging and record keeping systems consider all of the relevant events, triggers and information. However, standards can define a minimum set of generally applicable information elements. Besides these essential information elements, standards should set clear requirements on how to establish a proper logging plan and how to implement, test and document it, including the key criteria that AI system providers have to consider.
- **Transparency**: a broad range of information should be provided, supporting the understanding of how the AI system works, its characteristics, capabilities, strengths, limitations and performance, in line with the legal context. In particular, information about the potential circumstances that could give rise to risks, as well as information to support understanding the outputs of the system, enabling deployers to make informed decisions and take corrective actions if needed. Especially for high-risk AI systems, they should ensure that the transparency related information is comprehensive, meaningful, accessible and understandable for its intended audience.
- **Human Oversight**: assurance that AI systems stay within intended operational constraints, and that natural persons are able to control and override their outputs if necessary, must be established. Providers of high-risk AI systems should be able to translate the requirements defined in standards into concrete oversight measures, from the wide range of available options, based on the specific AI systems and the risks identified. Standards should also define a set of clear requirements on how the selection of those oversight measures has to be carried out, and how these have to be implemented and tested.
- **Accuracy**: the requirements that support providers of high-risk AI systems in the selection of relevant and effective accuracy metrics and thresholds must be defined. In addition, standards should define the processes and techniques that should be used to measure accuracy reliably, and to report it following best practices. These requirements should be especially clear for providers of high-risk AI systems, explicitly defining the criteria and priorities to be followed when implementing accuracy measurement methods, when choosing between metrics, thresholds and benchmarks, and when documenting the necessary evidence at the right level of granularity.
- **Robustness**: requirements related to the resilience of high-risk AI systems when deployed, including when facing errors, faults or inconsistencies in the environment of operation, should be defined. Providers of AI systems have at their disposal a broad range of technical and organisational measures that can support robustness and prevent harmful or undesirable behaviours. Standards are expected to support AI providers in defining those metrics in line with the intended use and potential risks of specific systems, as well as in implementing effective testing protocols.
- **Cybersecurity**: standards should describe technical and organisational measures to achieve a level of cybersecurity that is appropriate for the risks of AI systems. More specifically, the essential requirements for the implementation of a security risk assessment and mitigation plan for high-risk AI systems should be defined. Even if concrete security measures cannot be mandated in advance for every AI system, standards can define specific security objectives to be achieved (primarily at the system level), and how these should be verified through testing.
- **Quality Management**: quality management system (QMS) standards (similarly with risk

management) for existing product safety legislation are a useful reference, supporting compatibility with existing processes in some sectors. However, additional considerations are required to cover the specificities of AI products, whether embedded in physical products or in the form of software services. New standardisation on QMS should adopt a targeted focus on the specific risks addressed by the Regulation, as well as a product-centric view.

- **Conformity Assessment**: the procedures and processes required to assess conformity of high-risk AI systems, with the regulation prior to their placement on the market or when being put into service, should be defined. Additionally, the criteria for assessing the competence of the entities tasked with the conformity assessment activities should be also defined. Therefore, standardisation work for AI conformity assessment can be highly focused and precise, defining in practical terms how these procedures, processes and frameworks should be applied and adapted in particular for high-risk AI systems. In this context, the alignment between standards for conformity assessment and the various high-risk AI systems' requirements is essential and the coordination between parallel standardisation items should ensure that the resulting standards are complementary and fit their purpose.

Given the novelty of the AI Act and the complexity of its application due to the large number of involved stakeholders, even though the regulation is an industry-agnostic horizontal regulation, the European Commission has considered in the standardisation request additional vertical specifications regarding certain sectors[32]. In addition, algorithmic hiring cannot be classified in only one of the use case categories originally defined in the CEN-CENELEC roadmap[33]. Therefore, we highlight the need for specific standardisation actions for algorithmic hiring to ensure the compliance of the developed and used systems with the AI Act. Furthermore, we propose specific contributions to those actions.

## 2.3. The FINDHR case study

To detect and mitigate bias and discrimination in the multiple stages of algorithmic hiring, FINDHR[34] analysed all the elements of an AI recruitment system, as well as the needs of the multiple stakeholders interacting with such systems to identify the most adequate solutions for each entity and phase in AI lifecycle, including development and post-deployment.

An overarching approach designed for the algorithmic hiring field was proposed, with contributions that stressed responsible design from very early stages (i.e., how job roles are described and advertised) all along the chain, which comprises training, evaluation, deployment, and post-hiring impact assessment. During the development stage of AI systems, these solutions primarily include fairness-aware rankings and synthetic datasets. Throughout the entire AI system's lifecycle, FINDHR adapted an impact assessment and auditing framework, proposed a software development guide and explainability models, taking into account the perspectives of developers, recruiters and job seekers. In the post-deployment stage, a protocol for monitoring the operations of AI-based ranking systems in real-world applications based on secure third-party computation, was included. The guidelines, software, methodologies and other materials developed (Impact Assessment and Auditing Framework[35], Equality Monitoring Protocol[36], Software Development Guide[37], Fairness API[38], Synthetic CVs Dataset[39], Toolkits[40]) can be freely accessed through the project's web page. By wrapping these methodologies and results into guidelines with an appropriate structure, we aim to promote standardisation and best practices in algorithmic hiring, under the AI Act.

---

[32] Kilian, R., Jäck, L., & Ebel, D. (2025). European AI Standards-Technical Standardisation and Implementation Challenges under the EU AI Act. Available at SSRN 5155591

[33] CEN-CENELEC Focus Group Report: Road Map on Artificial Intelligence (AI)

[34] https://findhr.eu/

[35] https://findhr.eu/impact-assessment-and-auditing-framework/

[36] https://findhr.eu/equality-monitoring-protocol/

[37] https://findhr.eu/software-development-guide/

[38] https://findhr.eu/fairnessapi/

[39] https://findhr.eu/dataset-synthetic-cvs/

[40] https://findhr.eu/toolkits/

## 3. Recommendations for AI Act compliant standardisation actions

This chapter presents our recommendations for each of the areas specified by the EU Commission in its standardisation request (described in Section 2.2), focusing especially on the specific needs and requests by CEN-CENELEC, given the current lack of dedicated standardisation frameworks, or detailed guidelines.

In cases where JTC21 does not provide a clear (bullet-style or bold-style) structure for a specific area of AI, the authors derived and framed the additional needs from the surrounding context to help readers better understand the issue addressed. To ensure clarity and avoid misunderstandings, the recommendations that have an explicit and clearly defined bullet description in the JTC21's documentation are directly titled in this paper; the others (the ones identified by the authors) are marked with an asterisk (*). This approach is intended to make it easier for readers to locate, reference, and correlate the recommendations with the original JTC21 material. Furthermore, for every recommendation provided, we specify the type of change proposed, namely general (GE), technical (TE), or editorial (ED), in alignment with the template used in the CEN-CENELEC recommendation reports.

Although the recommendations we present were informed by FINDHR's outputs, they are not limited to its technical artefacts. They are formulated as domain-level standardisation needs for algorithmic hiring systems, especially ranking-based recruitment systems, and may be implemented using alternative methods, tools, or governance mechanisms. Following the developed approach for bias mitigation in algorithmic hiring systems, we propose an end-to-end debiasing pipeline, as shown in Figure 1, and present specific standardisation recommendations for the multiple processes included.

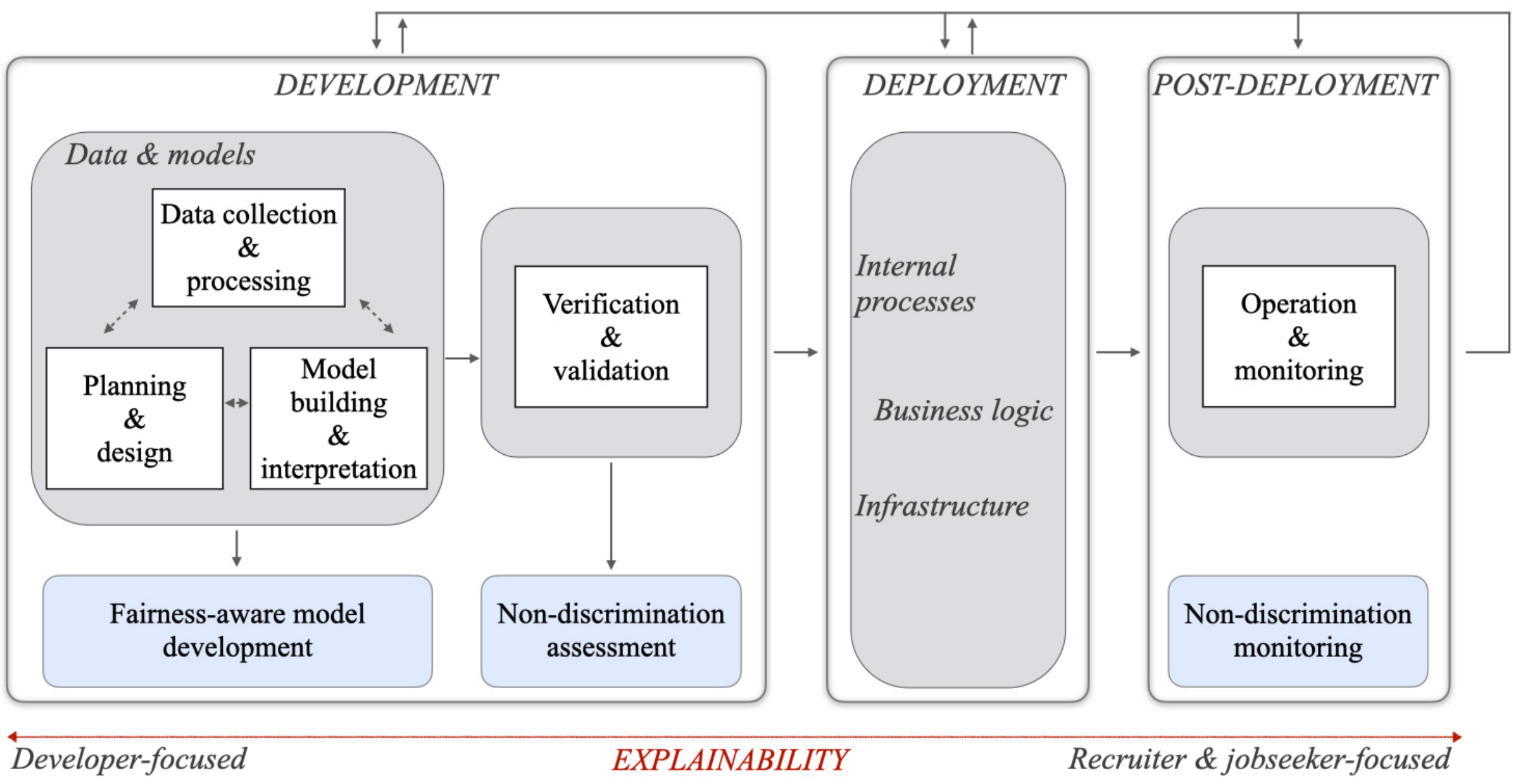

*Figure 1: High-level overview of the recommended pipeline and information flows, to counter discrimination in algorithmic hiring.*

Hence, our recommendations stand at the joint crossroads of technical, legal, and ethical implications, aiming to offer a practical and implementable approach to ensuring safe and lawful AI in employment. These recommendations were derived through the following four-step process:

I. Identification of the AI Act standardisation areas
II. Review of relevant horizontal AI standards and policy frameworks
III. Mapping of algorithmic hiring risks and FINDHR outputs to these areas
IV. Formulation of standardisation recommendations using the GE/TE/ED classification

## 3.1 Risk Management

The AI Act, particularly Article 9, considers risk management as a process planned and run throughout the entire lifecycle of a high-risk AI system, requiring regular systematic review and updating. FINDHR catered to fundamental rights' considerations in a systematic and iterative way and treated bias in AI hiring systems as an inevitable risk to the non-discrimination right. Among the crucial elements of this integration, the development cycle of the system should include a bias impact assessment of AI systems, called Fairness Impact Assessment (FIA), which supports developers, AI providers and other actors, in identifying and mitigating potential harms. These assessments were modelled to be carried out in a very systematic way for the purpose of identifying and evaluating potential discriminatory outcomes in the process of making algorithmic decisions. Moreover, in order to perform tests with fairness benchmarks, before deploying the system in production, the use of synthetic datasets –created from donated, anonymised CVs– is recommended. This approach gives the opportunity to evaluate, detect and disclose indirect discrimination risks in time.

Aligned with the categories described in the JRC Technical Report, we highlight that an AI-Act-oriented risk assessment is required to explicitly detect and evaluate both foreseeable and non-foreseeable risks, in terms of their potential impact on fundamental rights, health, and safety. As current ISO/IEC risk management standards (such as ISO/IEC 23894 and ISO 31000) mainly define risk in terms of organisational goals and system performance, they do not fully answer the needs of Article 9. As a result, these standards do not effectively implement the AI Act's human-centred notion of risk, including its demands for continuous monitoring and for assessing the acceptability of the risk.

Four key needs related to risk management, that systems for algorithmic hiring, and other high-risk applications, should cover, have been identified. Furthermore, to target each of those needs, we recommend that standards:

***Need 1 – Risk assessment***

I. Should include a mandatory, lifecycle-based risk identification focused on harms to individuals and society (for example as in the guidelines of FINDHR's Impact assessment and auditing framework).

II. Should use system cards as a tool to answer the AI Act Article 9 requirements on identifying and analysing foreseeable discriminatory risks and managing remaining risks. The risk assessment phase should explicitly require identification and analysis of foreseeable risks to fundamental rights (in particular discrimination risks), health, as well as risk to safety during all stages of the AI lifecycle, such as data preparation, model development, deployment, and post-market use. System cards take this approach a step further by including information about model choices, governance arrangements, and human involvement in the system's operation.

III. Should define Audit Operations (AuditOps) as a continuous lifecycle-based process that acknowledges that risks may either surface or evolve even during post-deployment and therefore must be reassessed in the course of repeated iterations. The need for continuous risk assessment, instead of a single ex-ante assessment, is also reflected in the AI Act's obligation to evaluate other risks that may arise on an ongoing basis, based on data gathered through the post-market monitoring system referred to in Article 72. While risk management under the AI Act is intended to cover all relevant categories of risk, AuditOps provides a detailed and operationalised framework for the continuous identification, documentation, and reassessment of discrimination and bias risks in algorithmic hiring systems. The concept of "Lifecycle and iterative risk assessment" is not new. ISO 31000 describes risk management as iterative, continuous and applicable across the entire life cycle. The same applies to ISO/IEC 42001, that introduces the Plan-Do-Check-Act (PDCA), which implies iteration and lifecycle coverage. However, while these standards address risk management in a horizontal and broader manner, in cases like (and not limited to) bias risks in algorithmic hiring, domain-specific operational guidance is necessary.

IV. Should combine pre-deployment testing, post-deployment monitoring, and iterative reassessment to detect, document, and mitigate discriminatory and bias-related risks throughout the AI system lifecycle. In ISO terms, a new iteration is what changes the organisational context, or is reviewed in relation to management and deviations in performance. In contradiction, according to FINDHR's outcomes, from such an implementation the "risks may emerge or evolve after deployment and therefore must be reassessed iteratively." In this sense, a different trigger logic arises (as shown in Table 1), in which observed impacts, drift, and emerging bias are also considered, while monitoring risks to individuals and groups. During runtime, this monitoring approach functions both as an ongoing monitoring and a mechanism for risk discovery, helping to address the requirements of the AI Act.

| Aspect | ISO standards | Proposed AuditOps |
|---|---|---|
| Trigger for reassessment | Organisational review cycles, audits | Observed impacts, drift, emerging bias |
| Focus of reassessment | Risk to objectives/system | Risk to individuals and groups |
| Post-deployment role | Monitoring as assurance | Monitoring as risk discovery |

*Table 1: ISO vs Proposed approach.*

V. Should require post-market bias monitoring as part of risk management, including the use of measurable fairness indicators and privacy-preserving techniques, in order to identify non-foreseeable and re-emerging discriminatory risks in line with Article 9 of the AI Act. Existing ISO standards address bias at different stages of the AI lifecycle and they do not require bias to be treated as a continuously monitored risk.

| Id | Clause/subclause | Change type | Comments | Proposed change(s) |
|---|---|---|---|---|
| RM1 | Risk management – general requirements | GE | Current international standards treat lifecycle and iterative risk management mainly as a managerial or organisational procedure. Yet, they do not demand that reassessment must be initiated in response to harms or harmful impacts on individuals observed during system operation, as mandated by Article 9 of the AI Act. | Include a requirement specifying that risk management iteration should be triggered by detected harms or negative impacts on individuals or groups, identified through operational (post-deployment) monitoring, and not solely by predefined management review cycles. |
| RM2 | Risk identification and assessment | GE | Existing standards generally fail to explicitly cover how to identify new, unforeseeable risks that arise after deployment, even though this is a central obligation under the AI Act. Monitoring is largely viewed as a control or assurance function, rather than as a mechanism for uncovering additional risks. | Introduce an explicit requirement that post-deployment monitoring should be used as a mechanism for identifying non-foreseeable risks, and that such risks shall be incorporated into the risk assessment process when detected. |
| RM3 | Risk management process | GE | International standards do not define a binding linkage between monitoring activities, risk reassessment, and system revision. As a result, detected issues may not systematically lead to updates of the risk management system. | Establish a mandatory connection between monitoring outcomes, risk reassessment, and system updates, so that any identified harms or new risks automatically initiate a reassessment and, when appropriate, lead to revisions of mitigation measures and system design (like AuditOps). |
| RM4 | Risk definition and scope | GE | Most of the current standards see risk in the context of the goals of the system/organisation, while the AI Act considers risk as the potential harm to individuals, their rights, as well as their health and safety. This difference is not clearly operationalized. | Explain, as far as risk management in AI systems is concerned, that risk should be understood and evaluated from the perspective of the potential harm that may be caused to people and society as a whole. Therefore such a human-centered understanding of risk should be applied to and operationalized in all activities concerned with risk identification, tracking, and continual updating of the risk assessment, as a matter of routine. |

*Table 2: Recommendations for Risk management (need 1).*

***Need 2 — Risk Treatment***

Building on the impact-focused risk assessment outlined above, the second requirement relates to Risk Treatment—that is, how the identified risks to individuals and groups are, in practice, removed, reduced, or otherwise managed. While existing ISO/IEC standards describe risk treatment at a high level, they do not sufficiently specify the AI-Act-oriented and impact-driven mitigation measures, nor do they require that such measures remain effective throughout system operation. Based on the gaps identified the following recommendations are proposed:

I. Since current standards define risk treatment mainly as a general choice of controls, without demanding a clear and explicit link between the identified harms to individuals and the mitigation actions taken, a direct correspondence between the specific harms and discriminatory effects (e.g., gender or ethnic minority background discrimination in hiring algorithms, age-based filtering in recruitment systems, etc.) revealed during the risk assessment should be established. This is intended to secure traceability between the observed impacts on individuals, or groups, and the selected mitigation measures.

II. The new standards should require that discriminatory risks are addressed through explicit bias mitigation measures applied at appropriate stages of the system's lifecycle (e.g. data, model, output), and that these measures are selected based on the type of harm identified.

III. Risk treatment should be accompanied by requirements for continuous verification of mitigation effectiveness, using operational monitoring results to determine whether risks have been sufficiently reduced or whether additional measures are needed.

IV. Risk treatment should be embedded in routine system maintenance and update processes, so that mitigation measures are adapted in step with the system whenever its context, data, or patterns of use change.

| Id | Clause/subclause | Change type | Comments | Proposed change(s) |
|---|---|---|---|---|
| RM5 | Risk management – risk treatment requirements | GE | Existing standards describe risk treatment at a high level but do not require mitigation measures to be explicitly linked to the specific harms or discriminatory impacts identified during risk assessment. As a result, risk treatment may remain generic and disconnected from the actual impacts on individuals and groups. | Add a requirement specifying that risk treatment measures should directly correspond to specific harms and discriminatory impacts identified during risk assessment, ensuring traceability between detected impacts and selected mitigation measures. |
| RM6 | Risk treatment – bias mitigation | GE | While bias is recognised as a risk source, existing standards do not require bias mitigation to be treated as a core risk treatment activity, nor do they specify that mitigation measures should be applied at appropriate stages of the system lifecycle depending on the type of harm identified. | Introduce a requirement that discriminatory risks should be addressed through explicit bias mitigation measures applied at appropriate stages of the AI system lifecycle (e.g. data, model, output), and that such measures should be selected based on the type and source of harm identified. |

| RM7 | Risk treatment effectiveness | GE | Current standards do not require providers to verify, over time, whether implemented risk treatment measures remain effective in reducing harms to individuals, nor do they link mitigation effectiveness to operational monitoring outcomes. | Require risk treatment to be paired with ongoing verification of how effective the mitigation measures are, relying on operational monitoring results to assess whether risks have been reduced to an acceptable level or if further or different measures are required. |
|---|---|---|---|---|
| RM8 | Risk treatment and system lifecycle | GE | Existing standards do not establish a binding connection between risk treatment measures and system maintenance or revision processes, potentially leading to static mitigation approaches that do not evolve when system behaviour or context changes. | Introduce a requirement for risk treatment to be operationally integrated into systems' maintenance and revision activities, ensuring that mitigation measures are updated when data, usage patterns, or deployment contexts change. |

*Table 3: Recommendations for Risk management (need 2).*

***Need 3 — Testing***

FINDHR proposed targeted testing practices that can be considered as risk-oriented testing in specific contexts. This approach was demonstrated through the implementation of fairness and reliability metrics and techniques for testing fairness effects within workflows, which rely on quantitative fairness metrics to evaluate whether bias mitigation measures are effective or not. These metrics serve to evaluate outcomes both before and after mitigation, enabling confirmation that discriminatory effects have been diminished. Therefore, the new standards should be designed to permit and facilitate risk-based testing of mitigation strategies using quantifiable fairness metrics, specifically to confirm whether the implemented controls actually reduce the identified discriminatory risks.

| Id | Clause/subclause | Change type | Comments | Proposed change(s) |
|---|---|---|---|---|
| RM9 | Testing of risk management measures | TE | Existing standards do not sufficiently address risk-oriented testing of mitigation measures, particularly in relation to verifying whether applied controls effectively reduce discriminatory harms. Testing is often framed generically and is not explicitly linked to the evaluation of bias mitigation effectiveness. | Introduce a requirement that standards allow and support risk-oriented testing of mitigation measures using measurable fairness indicators, specifically to verify whether applied controls reduce identified discriminatory risks. |
| RM10 | Testing of mitigation effectiveness | TE | Current standards do not explicitly require testing outcomes before and after the application of bias mitigation measures, limiting the ability to verify whether discriminatory risks have been effectively reduced. | Require that testing of bias mitigation strategies involves a comparative analysis of system results before and after mitigation, relying on quantitative fairness metrics to evaluate the extent to which discriminatory impacts are reduced. |
| RM11 | Testing within system workflows | TE | Testing is often treated as an isolated activity rather than being integrated into system workflows, which limits its relevance for assessing real-world impacts of mitigation measures. | Should be clear that testing of risk mitigation measures should be conducted within dedicated workflows or in environments that closely resemble production, so that the effectiveness of the controls can be evaluated under realistic operating conditions. |

*Table 4: Recommendations for Risk management (need 3).*

***Need 4 — Risk management system***

Risk management should be continuous, lifecycle-based, documented, and operationally enforced, rather than remaining a static governance artefact. Therefore, the following recommendations are proposed:

I. New standards should require that the risk management system explicitly covers all stages of the AI system lifecycle, including development, deployment, and post-market operation, ensuring continuity of risk management activities over time. The system's assessment should be an end-to-end process, covering phases from user/data acquisition and workflow definition through system feedback, since discrimination may still occur once the system is in use, and for this reason post-deployment monitoring is also required.

II. The risk management system should be seamlessly embedded into both system governance and daily operations, so that risk-related decisions can directly initiate corrective measures, system updates, and organisational actions while the system is running. Questions such as "Are systems in place to monitor the system?" and "What is the process for revising the algorithm?" should be clearly replied.

| Id | Clause/subclause | Change type | Comments | Proposed change(s) |
|---|---|---|---|---|
| RM12 | Risk management system – scope | GE | Existing standards describe risk management systems primarily at an organisational level and do not explicitly require coverage of the full AI system lifecycle, including post-deployment operation). | Include a requirement stating that the risk management system must encompass every phase of the AI system's lifecycle, including its development, post-(deployment and market) operation. |

| RM13 | Risk management system – continuity | GE | Current standards do not operationalise the requirement for risk management systems to function continuously and iteratively based on system operation and monitoring outcomes. | Require the risk management system to function as an ongoing, iterative process that uses monitoring and audit findings to periodically revise risk assessments and adjust mitigation measures over time. |
|---|---|---|---|---|
| RM14 | Risk management system – documentation | TE | Existing standards do not sufficiently require traceable documentation linking identified risks, mitigation measures, monitoring results, and system revisions; . | Add a requirement that the risk management system maintain structured records that ensure clear traceability between identified risks, their mitigation measures, monitoring results, and any resulting system modifications. |

*Table 5: Recommendations for Risk management (need 4).*

## 3.2 Data governance and quality

Ensuring the quality of training, validation, and testing datasets is a key requirement under the AI Act for high-risk AI systems (Article 10). This includes identifying and mitigating biases that could affect health and safety of persons, negatively impact fundamental rights, or lead to discrimination.

Furthermore, the AI Act identifies synthetic data, which mimics real-world patterns, as a promising way to detect bias in training, validation and testing data sets without using real personal data. Based on the Article 10(5) of the AI Act, the use of real sensitive data for bias detection is permitted only when strictly necessary and when no alternatives, such as synthetic or anonymised data, are effective. Therefore, synthetic data can play an important role in helping developers meet data quality obligations and fight against discrimination, as well as reduce reliance on sensitive real-world data. Although the ISO/IEC 5259 series offers a robust and comprehensive framework for managing data quality, it primarily defines data quality with respect to its usefulness for the organisation and its value for analysis. As the JRC has pointed out[41] this focus leads to weaknesses when the framework is applied in context of the AI Act's risk-based, human-centred approach, especially in relation to discrimination, fundamental rights, and proportionality.

Therefore, our guidelines, supplement the ISO/IEC 5259 by adding operational requirements that are specific to the AI Act. These guidelines define the fairness-critical data governance aspects (mainly from Bias and discrimination risk factors, Data representativeness and coverage, Data provenance and traceability, Feature selection and proxy detection, Human oversight touchpoints, Documentation and record-keeping obligations) that require prioritisation, provide justification for the necessity of specific data characteristics in the assessment of each individual risk, and specify how the data governance model decisions should be documented in a manner that is not only proportional but also open to enforcement.

***Need 1 — Data governance across the data lifecycle***

ISO/IEC 5259-3 and 5259-4 already address dataset design, collection, preparation, processing, and labelling throughout the lifecycle. Nonetheless, as highlighted by the JRC, the scale and complexity of the required processes may slow down achieving proportionate compliance with the AI Act. In addition, these standards do not clearly mandate that lifecycle decisions be tied to the risks faced by individuals.

[41] AI Watch: Artificial Intelligence Standardisation Landscape Update - Analysis of IEEE standards in the context of the European AI Regulation https://publications.jrc.ec.europa.eu/repository/bitstream/JRC131155/JRC131155_01.pdf

For high-risk AI systems, data governance throughout the entire lifecycle should be selectively delineated and rigorously justified with respect to its relevance under the AI Act. In particular, we recommend the:

- Lifecycle documentation explicitly identifies which dataset design and processing steps are critical for AI Act compliance, rather than requiring exhaustive documentation of all steps.
- Explicit documentation of assumptions about what the data represents, including which populations, behaviours, or social contexts are reflected and which are not.
- Identification of data gaps and shortcomings specifically in relation to risks to individuals and groups (e.g., under-representation or absence of certain demographic groups, missing or coarse-grained attributes relevant to fairness analysis, lack of ground-truth outcome labels for specific populations, or insufficient coverage of relevant social or occupational contexts).

| Id | Clause/subclause | Change type | Comments | Proposed change(s) |
|---|---|---|---|---|
| DG1 | Data governance – lifecycle coverage | GE | ISO/IEC 5259 offers a comprehensive definition of lifecycle governance, but it does not specify which aspects are essential for meeting AI Act requirements, resulting in substantial complexity for organisations trying to adopt it. | Ensure that data governance efforts are scoped proportionately by determining and recording which aspects of dataset design, collection, processing, and labelling are relevant to ensure data quality and bias mitigation, and therefore to reduce risks to individuals, in accordance with the Article 10 of the AI Act. |
| DG2 | Data assumptions | GE | Existing standards do not require explicit documentation of assumptions regarding what the data represents socially or demographically. | Introduce a requirement to document assumptions about population coverage, social context, and representativeness of training, validation, and operational datasets,, including known exclusions or limitations. |
| DG3 | Data gaps and shortcomings | GE | Data gaps are acknowledged but not explicitly framed as risks to individuals. | Require identification and documentation of data gaps and shortcomings specifically in terms of their potential impact on individuals and groups. |

*Table 6: Recommendations for Data governance and quality (need 1).*

***Need 2 — Assessment of data availability, quantity, and suitability***

Although ISO/IEC 5259 mainly addresses data availability and volume, it does so mostly from the technical sufficiency and system performance perspective. In contrast, Article 10 of the AI Act requires a basic assessment of whether the data are appropriate for the intended purpose, with particular attention to minimising discriminatory, biased, or otherwise harmful outcomes. Thereby widening the evaluation beyond technical appropriateness, it covers the need for safeguarding fundamental rights and societal interests. The following requirements are proposed:

- Assessment of data availability and quantity must consider whether protected or vulnerable groups are sufficiently represented;
- Suitability should be assessed in relation to the target deployment context, rather than solely with respect to the development environment;
- Insufficient representation of particular groups in the data should be regarded as a data governance risk, rather than just a data quality issue.

| Id | Clause/subclause | Change type | Comments | Proposed change(s) |
|---|---|---|---|---|
| DG4 | Data availability and quantity | GE | Current standards evaluate sufficiency primarily on technical grounds and do not explicitly address potential discrimination risks. | Require assessment of data availability and quantity in relation to the representation of individuals and groups potentially affected by an AI system. |
| DG5 | Data suitability | GE | Suitability is not clearly linked to the deployment context or to its effects on society. | Require that suitability assessments take into account the specific operational and social context in which the AI system will be used, including its potential effects on individuals. |

*Table 7: Recommendations for Data governance and quality (need 2).*

***Need 3 — Identification and treatment of unwanted bias in data***

Unwanted bias is currently treated in ISO/IEC 5259 as one attribute among many, without sufficient elaboration. Bias-focused standards (ISO/IEC 24027, TS 12791, IEEE 7003) provide valuable guidance, but they are not integrated into a data governance logic aligned with Article 10. Therefore, we propose that harmonised standards should incorporate requirements for:

- Explicit treatment of unwanted bias as a core data governance concern, not an optional quality check;
- Systematic identification of bias sources, including proxy attributes and data originating from external sources;
- Mandatory documentation on how identified biases are addressed through data design, preprocessing, or controlled use of synthetic data.

| Id | Clause/subclause | Change type | Comments | Proposed change(s) |
|---|---|---|---|---|
| DG6 | Data bias identification | TE | ISO/IEC 5259 considers bias primarily in broad, high-level terms and lacks concrete, practical instructions that are clearly mapped to the risk categories and mitigation obligations defined in the AI Act. | Require systematic identification of unwanted bias in datasets, including proxy attributes and external data sources, as part of data governance. |

| DG7 | Bias treatment measures | TE | Current standards do not mandate an explicit association between identified data biases and the corresponding mitigation measures. | Introduce a requirement that identified data biases be explicitly linked to documented mitigation measures within the data lifecycle. |
|---|---|---|---|---|

*Table 8: Recommendations for Data governance and quality (need 3).*

***Need 4 — Risk-oriented data quality attributes***

ISO/IEC 5259-2 provides a complete catalogue of data quality attributes, but defines data quality as organisational requirements. This definition should be re-oriented towards risks to individuals, including discriminatory risks, under the EU regulatory context. Based on the research on synthetic data (CVs) for fairness benchmarking held in FINDHR, we recommend the:

- Prioritisation of data quality attributes such as relevance, representativeness, correctness, completeness, and statistical properties based on their role in mitigating risks to individuals, in particular discriminatory risks
- Explicit justification of why certain attributes are prioritised for a given system and deployment context
- Assessment of statistical properties with respect to group-level impacts, not only aggregate performance

| Id | Clause/subclause | Change type | Comments | Proposed change(s) |
|---|---|---|---|---|
| DG8 | Definition of data quality | GE | Data quality is defined from an organisational perspective, which is insufficient for AI Act adoption. | Require data quality to be defined and assessed in terms of its potential role in creating or mitigating risks to individuals and groups, in particular discriminatory risks. |
| DG9 | Selection of data quality attributes | GE | Existing standards do not guide attribute selection based on European regulatory priorities. | Require prioritisation and assessment of data quality attributes based on their relevance to discrimination and fundamental rights risks in the intended deployment context. |
| DG10 | Statistical properties | GE | Statistical evaluation is not explicitly linked to group-level impacts. | Require assessment of statistical properties with attention to their effects on different groups potentially affected by the AI system. |

| DG11 | Bias detection legal basis | GE | The ISO standards assume data availability but do not address the GDPR's prohibition on processing sensitive data for bias and discrimination testing. | It should be mandatory to clearly define the legal foundation and guarantees for the processing of personal and, if necessary, sensitive data for bias and discrimination testing in accordance with Article 10 (data governance) and Article 9 (risk management) of the AI Act, as well as the GDPR.<br><br>*Note:* The Digital Omnibus proposal introduces additional clarifications regarding safeguards for processing sensitive data for AI risk monitoring (Article 4a). In particular, it points toward more explicit requirements on documenting security measures and pseudonymisation when such data are processed for risk and bias monitoring purposes. |
|---|---|---|---|---|

*Table 9: Recommendations for Data governance and quality (need 4).*

***Need 5 — Domain-specific synthetic data governance for fairness benchmarking***

Even though ISO/IEC 5259 recognises bias, representativeness, and identifiability, it does not define synthetic or semi-synthetic datasets as structured governance tools under AI Act constraints, therefore we contribute with a concrete operational pattern recommending:

- The use of semi-synthetic fairness benchmark datasets for pre-deployment testing of ranking-based hiring systems
- Alignment of synthetic datasets with both donor CV statistics and external workforce statistics
- Explicit limitation: synthetic datasets should not be used for post-deployment bias capture, acknowledging their inability to reflect evolving real-world dynamics
- Governance safeguards, including donor anonymisation, NDA-restricted access, relational splitting to reduce re-identification risk, and abort-generation rules where sample sizes are insufficient

| Id | Clause/subclause | Change type | Comments | Proposed change(s) |
|---|---|---|---|---|
| DG12 | Synthetic data for bias testing | GE | ISO/IEC 5259 does not provide normative patterns for synthetic data use under AI Act constraints. | Introduce synthetic-first bias testing as a recommended governance pattern for pre-deployment testing of high-risk hiring systems. |

| DG13 | Representativeness profiles | GE | Representativeness is required but not operationalised for workforce data. | Require domain-specific representativeness in profiles aligned with workforce statistics for hiring datasets. |
|---|---|---|---|---|
| DG14 | Synthetic dataset lifecycle limits | GE | Existing standards do not specify limitations on synthetic data reuse. | Limit the use of synthetic datasets strictly to pre-deployment fairness benchmarking purposes. |
| DG15 | Minimum-sample abort rule | TE | While the adequacy of the sample size is recognized, no explicit stopping (abort) criterion has been specified. | Require abortion of synthetic data generation where minimum sample thresholds for sector-demographic combinations are not met. |

*Table 10: Recommendations for Data governance and quality (need 5).*

## 3.3 Record keeping

A solid and well-defined record-keeping is not optional—it is the backbone of traceability and the key to powerful, real-world post-market monitoring of high-risk AI systems. Although current international frameworks offer a foundation for adequate record keeping, they do not answer the main concerns that are introduced by the AI Act about this topic. ISO/IEC 42001, for example, calls for automatic logging as part of its risk treatment measures; yet the accompanying guidance in its Annexes remains mostly generic and does not provide the detailed specificity needed to detect emerging risks or to conduct sophisticated fairness audits. Equally, IEEE 7001 offers valuable protocols for incident reconstruction and transparency—particularly aligned with Article 12—but its scope requires expansion to fully address broader conformity assessments. In detail, current standards do not resolve the conflict between the need to monitor AI systems for bias post-deployment and the GDPR's restrictions on processing sensitive data (e.g., ethnicity, religion, sexual orientation). Derived or inferred data is often incorrect and legally constrained and difficult to reconcile with GDPR compliance, while synthetic data is inadequate for post-market monitoring because it cannot capture the evolving real-world dynamics.

We propose a specific governance protocol that considers how monitoring-relevant data are collected, stored, protected, and accessed in a privacy-preserving manner to ensure comprehensive oversight and accountability. In detail, key aspects that should be included are the:

- Adoption of a Trusted Third Party (TTP) protocol, as a variation of Secure Multi-Party Computation (MPC) to collect sensitive data voluntarily from data subjects, strictly for monitoring purposes, with an active role of trusted third party. In detail, the TTP mechanism should specify an implementation pattern for logging and evidence handling that enables effective post-market monitoring. That would be a step forward to fulfil the requirements of Article 12 for privacy-preserving logging and evidence management and Article 72 for operational mechanism for post-market monitoring .
- Separation of duties in which neither the AI provider/deployer nor the trusted third party has access to the raw sensitive data of data subjects; instead, secure multiparty computation enables fairness monitoring of the AI system's outputs.
- Prohibition of using "proxies" or inferred sensitive data for monitoring without explicit consent or other legal basis.

| Id | Clause/subclause | Change type | Comments | Proposed change(s) |
|---|---|---|---|---|
| RK1 | Post-market monitoring plan | GE | According to Article 72 of the AI Act, the post-market monitoring system shall be based on a post-market monitoring plan, which shall be part of the technical documentation referred to in Annex IV. However, the Regulation neither provides details nor specifies a minimum internal structure for such plans. This is important, as it can ensure consistent coverage of risk-specific monitoring objectives and impacts on individuals and groups. | Post-market monitoring plans should follow a minimum standardized structure, including explicit sections covering risk-specific monitoring objectives, required indicators, triggers for re-assessment, and mechanisms for updating the plan. The structure should ensure that plans remain flexible but consistently address risks to individuals and groups.<br><br>*Note:* The Digital Omnibus proposal suggests removing the mandatory Commission template for post-market monitoring plans in order to increase flexibility. |
| RK2 | Post-market data collection | TE | Standards lack a privacy-preserving governance pattern for collecting sensitive data required for Article 10/Article 61 monitoring. | Recommend a Trusted Third Party (TTP) protocol, as a variation of Secure Multi-Party Computation (MPC) as the standard governance mechanism for processing sensitive data for post-market bias monitoring. |
| RK3 | Logging – ranking outputs & exposure information | TE | ISO/IEC 42001 and IEEE 7001 refer to logging only broadly (system events, failures) and do not require logging of AI output elements relevant to discrimination, such as ranking positions or exposure counts, which are essential for detecting bias. | Introduce a requirement that logging mechanisms should capture ranking outputs, exposure positions, and top-k visibility, thereby providing minimum evidence required to determine whether an AI system generates unequal outcomes once deployed. |

| RK4 | Logging of feature–protected attribute correlations | TE | Existing international standards do not mandate retention of feature-level distribution or inference traces, making retrospective identification of proxy-based discrimination impossible. | Require that logs record feature-level distributions and protected-attribute inference traces. That will enable the reconstruction of discriminatory pathways and will support the investigations into whether protected characteristics can be inferred or indirectly encoded in the future. |
|---|---|---|---|---|

*Table 11: Recommendations for Record Keeping.*

## 3.4 Transparency

As the EU non-discrimination law prohibits discrimination on the basis of ethnicity, gender, and other characteristics, the recruitment process should be transparent and fair towards candidates and other stakeholders involved in the recruitment processes. Especially in relation to candidates and the GDPR's rules, a hiring platform should inform candidates which categories of their personal data are used, for what purposes, and for how long the data are stored. Additionally, if a platform plans to perform automated decisions using AI systems, it is obliged to first make sure fully automated decisions are allowed and then ensure that candidates are informed about this and provided with a meaningful explanation of the logic behind the automated decisions they receive.

Furthermore, we highlighted the need for explainability of screening and ranking systems, supporting the different stakeholders involved in the recruitment process, the working logic of the systems and their outputs. Those aspects endorse the trust of the involved stakeholders, while permitting an improved testing and evaluation of the algorithms to assess whether they behave as expected, revealing vulnerabilities, detecting potential discrimination patterns while also supporting their compliance with the relevant legislation.

In order to design an explainable system supporting the different stakeholders involved in the recruitment process, two possible settings are plausible: the design of an explanation mechanism that takes as input a recommendation/ranking system and derive explanations to make the system's outcome transparent and explainable or to design a recommendation/ranking systems which are interpretable by-design meaning that the algorithm itself is able to provide reasons and motivations justifying its behaviour and recommendations. In both scenarios the explanations provided should be tailored to the involved stakeholders, reflecting differences in expectations, assumed knowledge, and required level of detail etc., and should be actionable, well-motivated, and fair. Furthermore, depending on the data domain over which an explanation has to be provided, the explanation may focus on the reasons for a decision (*factual explanations*) or the ways to change a decision (*counterfactual explanations*).

Despite partial coverage in ISO/IEC 42001, ISO/IEC 25059 and ISO/IEC TS 12792, current international standards do not sufficiently specify some aspects that are important for this specific topic, including:

- How transparency should be delivered (format, template, accessibility)
- How disclosures should be tailored to different stakeholders
- How training data provenance and representativeness should be presented
- What makes an explanation usable in practical hiring contexts (e.g., counterfactual advice)
- How transparency should be ensured on a continuous basis (transparency by design) across the AI lifecycle of the system/platform, including model versioning, retraining, and system maintenance

As a result, this creates a lack of alignment between the AI Act, which imposes legally binding obligations, and the standards, which remain high-level and, in some cases, optional. We aim to bridge this through the following recommendations.

| Id | Clause/subclause | Change type | Comments | Proposed change(s) |
|---|---|---|---|---|
| TR1 | Stakeholder-specific information duties | GE | Existing standards do not require transparency tailored to multiple user types; "users" are treated generically. | Introduce a requirement that transparency documentation should be tailored to stakeholder categories (roles), ensuring that candidates, recruiters, and auditors receive information aligned to their expertise and decision needs. |
| TR2 | Documentation of AI logic & decision-making | TE | ISO/IEC 42001 mentions documentation but not the obligation to describe decision logic. | Require AI systems to document their decision logic in clear terms, including which features are used, model type, and known limitations. |
| TR3 | Data transparency: training, testing & validation data | TE | Standards lack mandatory requirements for dataset provenance, representativeness, exclusions. | Introduce a requirement that documentation should include structured information on dataset composition, provenance, representativeness, exclusions, and known demographic gaps. |
| TR4 | Explainability – factual & counterfactual views | TE | No existing standard requires an actionable explanation ("what can I do to change the outcome", or guidance on how these explanations should be presented to users). | Require that explanation interfaces include both factual reasons for a given output and counterfactual guidance. |
| TR5 | Transparency format templates | TE | Standards provide taxonomies but no template formats, risking inconsistent documentation. | Introduce mandatory template formats for transparency documentation, analogous to structured cards (since these formats tend to be more user friendly for the final user). |
| TR6 | Accessibility & understandability rules | TE | No standards specify content readability or accessibility for users with different abilities. | Require transparency notices to take into account accessibility and clarity requirements (plain-language, multilingual, accessible formats). Accessibility constraints imposed by the EU should be also considered. |
| TR7 | Lifetime, maintenance & update transparency | GE | Standards do not propose how to notify users of model retraining or versioning. | Require that documentation specifies the expected lifetime and that users be informed whenever models, training data, or the operational context are modified. In addition, the documentation should describe the communication channel that will provide that information and the type of change. |
| TR8 | Transparency on automated decision use | GE | Standards do not require explicit notice of automated decisions or right-to-contest. | Require that systems provide clear notice when automated decisions are used and inform individuals of human-review and contestation rights. |

*Table 12: Recommendations for Transparency.*

### 3.5 Human oversight

Any AI system considered as high-risk must be subject to adequate human oversight to ensure it runs safely, fairly, and accountably. From the perspective of algorithmic hiring, high-risk AI systems should not produce or enforce decisions that affect candidates without meaningful human understanding, review, intervention, and remediation mechanisms throughout the system lifecycle.

Human oversight should be addressed through a End-to-End Socio-Technical Algorithmic Audit (E2EST/AA), which incorporates human input across pre-processing, in-processing, and post-processing phases. In particular, this methodology incorporates human input throughout the pre-, in-, and post- processing phases of algorithmic hiring systems. Any human auditor, working with tools such as System Cards and System Maps, can find themselves supported by thorough documentation and visualisations of data flows, decision logic, and risks. This support should allow them to question, interpret, and intervene with the system behaviour.

In addition, we propose an Impact assessment and auditing framework that scores bias-detection ability, reparative mechanisms at multiple stages, and interpretability, allowing auditors to evaluate whether systems treat individuals fairly, especially with sensitive decisions such as hiring. These are precisely the features envisioned in Article 14 of the AI Act, for which meaningful human oversight goes from being a possibility to an intrinsic part of the design and governance of algorithmic hiring tools. In this way, we propose a model for a replicable way to incorporate human oversight into AI systems used for recruitment and employment, and ensure that decisions affecting individuals — including hiring decisions— remain understandable, reviewable, correctable, and contestable. In the recruitment context, human oversight is essential as fully automated decisions may determine the employment strategy/decisions having significant consequences for individuals' rights, health, and livelihood, as well as legal consequences for a hiring company. That is one of the reasons why oversight should not be optional but rather a built-in responsibility. Human operators, such as recruiters, HR professionals, and hiring managers, must always be able to understand how a system works, how to detect, using a user-friendly documentation and explanation, errors or discrimination, as well as to be able to intervene when necessary. This also includes the ability, for recruiters and HR professionals, to verify or confirm system outputs before these take effect.

Current international standards offer a partial coverage (especially to the topic of -high risk-algorithmic hiring). ISO/IEC 42001 acknowledges oversight but does not define how it must be implemented. ISO/IEC TS 6254 lists XAI tools but does not specify when or whether they must be used, nor how explanations prevent automation bias. On the same logic the ISO/IEC TS 8200 introduces controllability, but does not specify user-roles, interaction-design requirements, or organisational safeguards, for the method to follow, to ensure oversight is not concrete and holistic. Finally, ISO/IEC 25059 mentions quality attributes like intervenability, but does not provide strict requirements.

***Need 1 – Effective Human Oversight through User Understanding and Training****

Human oversight in algorithmic hiring only performs well when the human operators involved in the process are able to understand and question the results and recommendations generated by the system. Users should be able to understand how the system works and why it behaves as it does. Furthermore, individuals interacting with the system must be able to interpret decisions, ask questions, and intervene when needed. Accordingly, well-structured training should ensure that human operators involved in the hiring process can recognise when the system is not performing as expected, including cases where bias appears or outcomes differ between groups. Without this capability, oversight may become an administrative step (adding a bureaucratic layer) rather than a real safeguard that is necessary for cases like hiring algorithms.

| Id | Clause/subclause | Change type | Comments | Proposed change(s) |
|---|---|---|---|---|
| HO1 | Training of the human overseeing the system | GE | Current standards fail to specify necessary skills or formal training for human overseers, and they also do not support preparation for automation bias, potential system failures, or performance gaps across different subgroups. | Introduce a requirement that human overseers should receive structured, scenario-based training covering system capabilities, limitations, subgroup performance, automation bias, and failure-mode recognition before they are authorised to operate or monitor the AI system. |
| HO2 | Human–AI interface usability enabling intervention | TE | ISO/IEC TS 8200 & ISO/IEC 25059 acknowledge controllability and interpretability, but provide no concrete requirements ensuring that interfaces enable comprehension, show uncertainty, or prevent passive approval of outputs. | Require human–AI interfaces to be designed in accordance with cognitive-ergonomics principles, visibly signal uncertainty (e.g., confidence scores), avoid binary outputs where active judgment is required, and present information in formats requiring human evaluation prior to decision finalisation. |
| HO3 | Organisational measures for oversight | TE | Current standards do not specify oversight roles, escalation paths, redundancy, or dual confirmation requirements; oversight often belongs to a single person without accountability or backup. | Introduce a requirement that oversight should include defined roles, responsibilities, escalation procedures, redundancy mechanisms, and, where appropriate, dual confirmation for decisions affecting individuals' rights before they take effect. |
| HO4 | Controllability / Intervenability of system operation | TE | Standards acknowledge controllability conceptually but do not require that systems allow pausing, reversing, or overriding decisions before consequences materialise. | Require that systems include mechanisms to pause, reverse, or override AI outputs prior to decision execution and shall prevent irreversible automated decisions without meaningful human review. |
| HO5 | Oversight-linked operational triggers | TE | Existing standards do not require monitoring triggers that actively alert humans when intervention may be needed (e.g., drift, loss of performance, bias spikes). | Introduce a requirement human oversight to be supported by operational activities that identify/reveal anomalies and require acknowledgement or corrective action by authorised personnel. |

*Table 13: Recommendations for Human oversight (need 1).*

***Need 2 – Appropriate design of the human–AI interfaces***

An appropriate design of the human–AI interface is critical to make oversight operational rather than symbolic. The interfaces must allow humans to understand, question, and act on systems' outputs before decisions are executed and/or responses are provided. As a ground rule, the systems used in hiring processes should present the logic, as well as the influencing factors, in a comprehensible way, to avoid passive binary outputs, signal uncertainty, and provide mechanisms for intervention. This approach should be highlighted in the standards in order for the users to be able to monitor, pause, or override automated outcomes.

| Id | Clause/subclause | Change type | Comments | Proposed change(s) |
|---|---|---|---|---|
| HO6 | Interface comprehension – information display | TE | Existing standards acknowledge interpretability but do not require that user interfaces present information in a format that a human can meaningfully understand and act on before decisions are made. | Introduce a requirement that interfaces should present the decision logic, the factors influencing ranking outcomes, and relevant context information in a comprehensible format that enables natural persons to evaluate and question the output prior to decision execution. |

| HO7 | Uncertainty and confidence signalling | TE | No AI-relevant standard currently mandates that user interfaces surface uncertainty indicators (confidence scores, lack of coverage warnings), which are essential to prevent automation bias. | Require that interfaces visibly signal system uncertainty (e.g., confidence levels, "data mismatch" warnings, flagging when model input is out-of-distribution), enabling humans to recognise when outputs may be unreliable. |
|---|---|---|---|---|
| HO8 | Cognitive-ergonomics & active-engagement design | TE | International standards do not require interfaces to prevent passive approval. Binary outputs (e.g., “hire/reject”) reinforce automation bias by not requiring human reasoning. | Require that interfaces avoid binary automated outputs where human judgment is required and instead provide multi-component information (e.g., factor breakdown, score distributions) that must be cognitively processed by a human before decisions are finalised. |
| HO9 | Action-oriented explanations (factual & counterfactual) | TE | Standards do not specify that explanations must be actionable or provide change-paths for individuals affected; this limits their usefulness for fairness and remedy. | Require that explanation mechanisms provide both factual (why this outcome) and counterfactual (what would need to change) information, tailored to the needs of distinct user groups (recruiters, candidates, auditors). |
| HO10 | Intervention-enabling UI design | TE | Standards on controllability (ISO/IEC TS 8200) define state-control concepts but do not require UI layouts that enable pausing, overriding, or reversing decisions before they take effect. | Introduce a requirement that human–AI interfaces should provide clearly accessible control mechanisms allowing authorised humans to pause evaluation, reverse automated decisions, or block a decision from being finalised without human determination. |
| HO11 | Stakeholder-specific UI tailoring | GE | Current standards do not specify tailoring for differentiated users (candidate vs recruiter vs auditor). Uniform interfaces do not satisfy rights-based needs. | Require that interfaces provide differentiated views and explanations in the depth appropriate to the role (candidate, recruiter, oversight authority, auditor), ensuring accessibility and clarity for each category. |

*Table 14: Recommendations for Human oversight (need 2).*

***Need 3 – Organisational measures***

Organisational measures are required to guarantee that control is not dependent on an individual but is implicitly incorporated in the governance of the system. Organisational safeguards should clearly specify roles, routes, responsibilities for review, and backup mechanisms, that all together will ensure timely detection of discriminatory risks. They should also ensure that accountable human oversight and the possibility of intervention are always in place, since oversight becomes meaningful only when organisations can trace *who* is responsible, *how* a decision can be challenged, and *what triggers* lead to system revision.

| Id | Clause/subclause | Change type | Comments | Proposed change(s) |
|---|---|---|---|---|
| HO12 | Role assignment & accountability | GE | Existing standards do not specify who, within an organisation, is accountable for monitoring, intervening, and triggering system revisions, leaving oversight unanchored. | Introduce a requirement that organisations should define formal oversight roles responsible for monitoring, intervention, and escalation, including named accountability and decision authority. |
| HO13 | Escalation & multi-stage governance | GE | Oversight today often stops at observation — no requirement ensures that detected risks automatically escalate into action or trigger revision. | Require that organisational oversight includes escalation procedures and structured workflows that prescribe actions when risk signals or discriminatory harms are detected. |
| HO14 | Redundancy and dual confirmation for consequential decisions | GE | Current standards do not mandate that high-impact decisions (e.g., applicant rejection) require more than one human or checkpoint, risking unjustified automated outcomes. | Introduce a requirement that decisions affecting individuals' employment rights should include redundancy or dual confirmation before outcomes are finalised, where proportionate. |
| HO15 | Operational integration of oversight | GE | Oversight mechanisms are frequently organisational add-ons rather than integrated into daily system operations or CI/CD update processes. | Ensure that oversight responsibilities, escalation pathways, and update procedures are built directly into day-to-day operations and system maintenance cycles, so they function in real time rather than as after-the-fact reviews. |

*Table 15: Recommendations for Human oversight (need 3).*

***Need 4 – Ensure Controllability and Intervenability***

Since high-risk AI in hiring must allow a human to stop, change, or reverse a system decision before it harms job candidates, human oversight is essential to control the process and timely intervene when the system operates out of the boundaries. Interfaces and workflows must also permit to pause a system, reject its results, or move a case to human review. When such control is missing or hard to reach, human oversight becomes critical, especially in high-risk cases such as hiring, where decisions can significantly affect individuals' lives. Therefore, the requirements, including the ability to pause the systems or to reverse their outcomes, should be explicitly implemented in the systems.

| Id | Clause/subclause | Change type | Comments | Proposed change(s) |
|---|---|---|---|---|
| HO16 | Ability to pause or stop AI actions | TE | Current standards describe control in theory but do not require systems to include a clear way for a person to pause or stop automated decisions. | Require that systems include an easy-to-use pause or stop control that a trained user can activate before a decision takes effect. |
| HO17 | Override or reverse decisions | TE | Standards do not require that a person can override or undo an AI decision before it impacts an applicant. | Introduce a requirement that systems allow a human to override or reverse automated decisions before they are applied to an individual. |
| HO18 | Visibility of control options | TE | Control features may exist but are not always visible or easy to find. A hidden control is not usable. | Require that control options (pause, override, escalate) appear in the main interface, in clear language, and without extra navigation steps. |
| HO19 | Control ownership | TE | Standards do not state who is allowed to use control functions. This creates confusion and weakens oversight. | Require that systems define which user roles can activate control functions and record who took the action. |

*Table 16: Recommendations for Human oversight (need 4).*

## 3.6 Accuracy

Under the AI Act, high-risk AI systems have to be technically accurate and robust with the aim to minimise errors and harmful results for individuals. In recruitment settings, this means ensuring that candidate assessments are fair and truly ascertain the candidate's qualification and suitability for a position. Accordingly, errors in classification or ranking may disadvantage some groups systematically and eventually result in discriminatory decisions.

Our study tackled accuracy through bias-aware evaluation protocols throughout the entire hiring pipeline. Specifically, we put forward a modular evaluation for input, output, and outcome fairness that allows developers and AI system providers to detect accuracy failures at various points in the Applicant Tracking System (ATS) pipeline. These points of failure may encompass inconsistencies in training data, imbalances in candidate exposure, and outcome disparities in final ranking. This approach also supports secure multi-party computations to calculate fairness metrics without infringing on sensitive personal data of job candidates, thus allowing for accurate and privacy-preserving evaluations of candidate outcomes.

The developed synthetic CVs dataset acts as a benchmarking tool and can be used to analyse the rank assignments contained in the hiring systems prior to deployment. The hiring organisations will be able to verify if their systems operate correctly under controlled, representative conditions. This way, it is ensured that the accurate measurement of the predictive performance of the system is comparable to so-called regulated accuracies, which contributes to trusted AI in the employment context.

Current ISO/IEC work gives only general guidance on accuracy (ISO/IEC 24029-1; 24029-2; DIS 25059) and does not specify how providers should select metrics, test accuracy over time, or justify thresholds relative to risks to individuals' characteristics that was shown to be important for this feature. We propose the use of  concrete methods, that should serve as standardisation input: bias-aware modular evaluation for input–output–outcome fairness, secure multi-party computation for privacy-preserving fairness testing, and semi-synthetic datasets to benchmark rank-output behaviour before deployment.

***Need 1 – Metrics and thresholds***

| Id | Clause/subclause | Change type | Comments | Proposed change(s) |
|---|---|---|---|---|
| AC1 | Metric justification | TE | Standards do not require providers to explain *why* they chose a metric or whether it reflects fairness or performance on the affected group or groups. | Require that metric selection is justified using controlled benchmark tests that reflect hiring-task behaviour. |
| AC2 | Threshold justification | TE | Thresholds can be set arbitrarily without evidence. | Require that threshold values be documented and justified using modular fairness/ accuracy testing at input-output-outcome level. |

*Table 17: Recommendations for Accuracy (need 1).*

***Need 2 – Design considerations***

| Id | Clause/subclause | Change type | Comments | Proposed change(s) |
|---|---|---|---|---|
| AC3 | Accuracy-critical design choices | TE | ISO standards do not show how design choices (text parsing, model scoring, ranking logic) influence accuracy downstream. | Require that providers document how design decisions affecting accuracy on applicants were made and validated (e.g., feature choice, ranking logic, matching rules). |

*Table 18: Recommendations for Accuracy (need 2).*

## 3.7 Quality management

Based on the AI Act, providers of high-risk AI systems are required to establish and maintain a Quality Management System (QMS) to ensure compliance with the legal obligations set out in this regulation across the entire life cycle of the AI system. Such processes should include design control, data management, testing and validation, documentation, corrective actions, and post-market monitoring. Given the sensitivity of personal data involved in algorithmic hiring and the potential system impact on access to the job market, quality management has become the highest priority for cleansed lawful operations as well as ethical and socially responsible operations of the algorithmic hiring systems.

We recommend that quality management is instantiated as a system of modular tools and documented procedures against the most critical phases during the development and deployment of AI-based ranking systems. These tools support providers in setting up repeatable, auditable workflows for testing and evidencing fairness, data governance, and performance validation. Introducing concrete evaluation criteria and control mechanisms for analysing input, output, and outcome fairness at varying levels of system decomposition, or including a continuous fairness monitoring layer after deployment, ensure that any undesirable discriminatory or performance drifts can be flagged in real time, making the systems compliant and trustworthy as they interact with changing user behaviours and labour market conditions. These aspects reflect the AI Act's requirement for quality control throughout the AI lifecycle. In particular, they correspond to Articles 17–19, that require providers to undertake systematic risk management, validate outputs, and keep technical documentation updated.

By providing a reusable methodological foundation and sectoral guidance, our suggestions aim to support not only the internal QMS for AI system providers, but also to better align with the future European harmonised standards under the AI Act. To effectively operationalise Article 17 of the AI Act in the recruitment domain, harmonised standards must provide concrete instructions on how a QMS should be implemented, evidenced and traced. This implies that quality management should be systematically designed and integrated across the AI lifecycle. We propose that the quality management requirement should be more operational in practice, through repeatable and auditable procedures, fairness and performance controls and continuous quality monitoring after deployment.

***Need 1 – Strengthening Justification of QMS Controls****

Regarding justification, the key question is whether documented justification for **decisions to adopt or not specific QMS controls**, should be required.

| Id | Clause/subclause | Change type | Comments | Proposed change(s) |
|---|---|---|---|---|
| QM1 | Justification of QMS controls | TE | ISO/IEC 42001 allows organisations to choose which controls to adopt, without specifying what is an acceptable justification/ reasoning. | Require a well formed justification template documenting: (i) *which* controls are included/ excluded, (ii) *reason*, (iii) *evidence*. |

*Table 19: Recommendations for Quality management (need 1).*

***Need 2 – Strengthening Post-Market Monitoring Requirements****

Regarding post-market monitoring of AI systems, an important consideration is the need for more detailed requirements that define what should be monitored during operation.

| Id | Clause/subclause | Change type | Comments | Proposed change(s) |
|---|---|---|---|---|
| QM2 | Content of post-market monitoring | TE | ISO/IEC 42001 only requires generic performance monitoring. | Require continuous monitoring of discriminatory impact signals (exposure drift, ranking outcome disparities, subgroup pattern changes). |
| QM3 | Tailored monitoring to risks missed earlier | GE | No mechanism to detect newly emerging risks. | Require the monitoring layer to detect impacts not identified during pre-deployment testing. |

*Table 20: Recommendations for Quality management (need 2).*

***Need 3 – Strengthening Documentation and Auditability for Quality Management****

Documentation needs are of vital importance, not only for the understanding of the code and the system but also for the auditability of each function and service, especially since AI algorithms are, in many cases, difficult to analyse. In particular, more clarity and consolidation is required.

| Id | Clause/subclause | Change type | Comments | Proposed change(s) |
|---|---|---|---|---|
| QM4 | Documentation consolidation | ED | 42001 documentation is distributed across many artefacts, making compliance difficult. | Require consolidation of core documentation — design decisions, monitoring evidence, risk register, modification log — into one structured QMS file. |

*Table 21: Recommendations for Quality management (need 3).*

## 3.8 Conformity assessment

The EU AI Act defines the process of conformity assessment for high-risk AI systems, including algorithmic hiring, to verify that they fulfil their regulatory obligations before being placed on the market or before being used in practice. The conformity assessment verifies that these systems are compliant not only with performance, accuracy, and standards, but also with legal requirements, including data protection, technical robustness, and ethics. The conformity assessment is thus a pre-condition for setting up the systems lawfully as well as a structured approach throughout the lifecycle for risk management and accountability.

The conformity assessment process under FINDHR was supported by designing an all-encompassing algorithmic hiring auditing framework that should be applied at any point along the full lifecycle of such systems, including tools and techniques for ex-ante validation (before deployment), continuous monitoring of performance and fairness (after deployment), and full documentation of procedures and results. In order to foster traceability and verifiability, certain practices for documentation that should be embedded in the technical file, as provided in Article 11 of the AI Act, were brought forth. For instance, bias metrics, fairness testing outcomes and synthetic benchmarking datasets, fortify the evidence towards proving a system's conformity. These instruments also help ensure ongoing conformity throughout the entire lifespan of the AI system, and should be aligned with Articles 17–19 of the AI Act.

Moreover, a design philosophy that heavily leans on modularity and adaptability, which makes tools suitable for both internal conformity assessments –that may be required in some cases (for instance, for systems developed in-house)– and external assessments, performed by the notified bodies in other cases is proposed. As part of conformity assessment under the AI Act, post-market monitoring (Article 72) should be performed, ensuring that any modification in the system’s behaviour after deployment is monitored and traced, concerning anything that might be unfair or discriminatory for the affected individual. In addition, FINDHR supported pre-deployment assessments and laid the foundation for continuous conformity, thus building trust and legal compliance in real-life hiring applications.

***Need 1 – Strengthening Conformity Assessment for Algorithmic Hiring****

Conformity assessment for algorithmic hiring requires concrete tools, evidence formats, and lifecycle-based triggers. Accordingly, the proposed additions to the future standards can be found in the following table.

| Id | Clause/subclause | Change type | Comments | Proposed change(s) |
|---|---|---|---|---|
| CA1 | Ex-ante validation before deployment | TE | AI Act Annex VI requires pre-market conformity but ISO CASCO toolbox does not introduce any AI-specific validation steps | High-risk AI hiring systems should undertake pre-deployment evaluation using structured lifecycle audits covering data → model → output → outcome fairness |

| CA2 | Fairness testing as mandatory evidence | TE | Current conformity frameworks do not require fairness evidence; bias can remain invisible | Measurable fairness indicators and bias-testing results should be part of the Technical File submitted for conformity (that could be also part of the documentation) |
|---|---|---|---|---|
| CA3 | Benchmark datasets for evidence | TE | Conformity bodies lack standard testing datasets → creates incomparable evidence | Sector-specific benchmark datasets should be used to validate ranking & exposure accuracy before deployment in a systematic manner. |
| CA4 | Documentation traceability | TE | Article 11 demands traceable documentation, but ISO 42001 lacks templates | Require a minimum standardized documentation template listing: system purpose → data → model → fairness tests → monitoring plan |

*Table 22: Recommendations for Conformity assessment (need 1).*

***Need 2 – Extending Conformity Assessment Requirements Beyond High-Risk AI****

Since conformity assessment is relevant for all AI systems and not only for those classified as high-risk, the following proposed changes introduce generic and technical clarifications on how internal and external conformity assessments should be operationalised in practice.

| Id | Clause/subclause | Change type | Comments | Proposed change(s) |
|---|---|---|---|---|
| CA5 | Internal conformity assessment | GE | AI Act Annex VI allows provider to carry out self-assessment but gives no guidance on how such assessment should be done in practice | Internal assessments should be based on lifecycle audit maps showing system phase → evidence → risks |
| CA6 | External conformity by CABs | TE | CABs need structured evidence packages | Require submission packages to include model cards, fairness logs, dataset sheets & monitoring plans as standard evidence |

*Table 23: Recommendations for Conformity assessment (need 2).*

## 4. Conclusion and Future Directions

AI standardisation is, by nature, a recurrent and iterative process that will continue to evolve as AI itself is being reshaped by new technological developments. The same reformation applies to high-risk AI applications and systems where best practices are more than vital and have to be standardised for compliance. In this paper, we have presented recommendations and best practices to support concrete standardisation actions for algorithmic hiring. Our study contributes to the emerging discussion on vertical standardisation by addressing a high-risk AI use case in the broader area of algorithmic hiring. Furthermore, many of those could be generalised to cover other domains with similar problems and characteristics.

Since the Digital Omnibus is expected to enter in force from August 2026, while the EN 18228 – AI Risk Management draft has been recently finalised, and given the unstable nature of negotiation processes, some of the requirements addressed in this work may be modified in short/mid term. However, through this work, we aim to initiate a broader discussion on vertical standardisation, recognising that each domain involves different constraints, requirements, motivations, and restrictions. The attempt presented in this paper, focused on the area of algorithmic hiring, highlights that standardisation cannot be limited to a specific entity, process, or algorithmic component. Instead, both the legislative context and the interconnectivity of software components within AI-enabled systems indicate the need for a holistic approach that addresses the full spectrum of the system. In this regard, the development of supportive tools, such as benchmark datasets and sandboxes, together with dedicated guidelines, should be further promoted in order to adequately address the rising concerns related to the use of AI in high-risk domains.

***Acknowledgement***

This paper was partially funded by the Horizon Europe project FINDHR (Fairness and Intersectional Non-Discrimination in Human Recommendation), grant agreement no.: 101070212. Our work was inspired by the research developed as part of this project by its consortium partners and collaborating experts, including: *Arora, P., Aszódi, N., Azores, M., Biega, A., Borghesi, D., Capasso, M., Carvajal, M., Castillo, C. Daviet, M., de Rijke, M., Dimopoulou, E., Escriba, G., He, C., Fabris, A., Galdon Clavell, G., Gonzalez, L., Graus, D., Mastropietro, A., Meijers, G., Monreale, A., Montesinos, S., Moudatsou, M., Mueller, A., Plackis-Cheng, P., Rotelli, D., Ruggieri, S., Rus, C., Saldivar, J., Via, A., Vorobyeva, T., Yates, A. and Zuiderveen Borgesius, F.* We would like to thank all of them.